\documentclass[conference]{IEEEtran}
\IEEEoverridecommandlockouts
\usepackage{cite}
\usepackage{amsmath,amssymb,amsfonts,commath}
\usepackage{algorithm}
\usepackage{algorithmic}
\usepackage{graphicx}
\usepackage{textcomp}
\usepackage{xcolor}
\usepackage{amsthm}

\usepackage{subfigure} 
\usepackage{hyperref}
\usepackage{url}
\usepackage{hyperref}
\usepackage{bm}

\newcommand\Tstrut{\rule{0pt}{2.ex}}         

\def\abovestrut#1{\rule[0in]{0in}{#1}\ignorespaces}
\def\belowstrut#1{\rule[-#1]{0in}{#1}\ignorespaces}

\def\abovespace{\abovestrut{0.20in}}

\def\belowspace{\belowstrut{0.10in}}

\DeclareMathOperator*{\argmin}{arg\,min}

\def\BibTeX{{\rm B\kern-.05em{\sc i\kern-.025em b}\kern-.08em
    T\kern-.1667em\lower.7ex\hbox{E}\kern-.125emX}}
\begin{document}

\title{Generating Compact Tree Ensembles via Annealing}

\author{\IEEEauthorblockN{Gitesh Dawer}
\IEEEauthorblockA{\textit{CoreML Group} \\
\textit{Apple Inc.}\\
Cupertino, California, USA \\
dawergitesh@gmail.com}
\and
\IEEEauthorblockN{Yangzi Guo}
\IEEEauthorblockA{\textit{Department of Mathematics} \\
\textit{Florida State University}\\
Tallahassee, Florida, USA \\
yguo@math.fsu.edu}
\and
\IEEEauthorblockN{Adrian Barbu}
\IEEEauthorblockA{\textit{Department of Statistics} \\
\textit{Florida State University}\\
Tallahassee, Florida, USA \\
abarbu@stat.fsu.edu}
}

\maketitle

\begin{abstract}
Tree ensembles are flexible predictive models that can capture relevant variables and to some extent their interactions in a compact and interpretable manner. Most algorithms for obtaining tree ensembles are based on versions of boosting or Random Forest.
Previous work showed that boosting algorithms exhibit a cyclic behavior of selecting the same tree again and again due to the way the loss is optimized. At the same time, Random Forest is not based on loss optimization and obtains a more complex and less interpretable model. In this paper we present a novel method for obtaining compact tree ensembles by growing a large pool of trees in parallel with many independent boosting threads and then selecting a small subset and updating their leaf weights by loss optimization. We allow for the trees in the initial pool to have different depths which further helps with generalization. Experiments on real datasets show that the obtained model has usually a smaller loss  than boosting, which is also reflected in a lower misclassification error on the test set.
\end{abstract}


%
\IEEEpeerreviewmaketitle

\section{Introduction}
In this work we are interested in finding parsimonious tree ensembles that minimize a loss function constrained to contain a small number of trees (tree sparsity) of any depth in a prescribed range.  As the space of decision trees is combinatorially complex, we are limited to suboptimal methods for loss minimization with  tree sparsity constraints.

One such method is provided by boosting, which adds one tree at each iteration to minimize a loss function in a greedy fashion. Different versions of boosting are aimed at minimizing different loss functions, but Gradient Boost \cite{friedman2001greedy} is a generic boosting algorithm that can be used for any differentiable loss function.

However, boosting algorithms have shown some difficulties in loss minimization when many boosting iterations are used. In such cases, a cyclic behavior has been observed \cite{rudin2004dynamics} where the same or a very similar weak learner is selected again and again at later boosting iterations.

One possible explanation for this behavior when the learners are decision trees comes from the correlation between the newly introduced tree and the already existing ones, which is reflected in the need to change their leaf weights for loss minimization. Since the leaf weights for the already existing trees are not updated, the only way to change them is to add another similar tree that makes up for the weight changes. 

In this work, we propose to find the trees and update their leaf weights simultaneously. Instead of using a forward-selection type of approach that is taken by boosting, where the ``best'' tree is added conditional on the already selected trees, we will use a stepwise approach that obtains a large pool of trees by boosting, then simultaneously selects a small number of them and updates their weights using the recently introduced Feature Selection with Annealing (FSA) \cite{barbu2016feature} algorithm.
\begin{figure*}[ht]
\centering
\includegraphics[width=14cm]{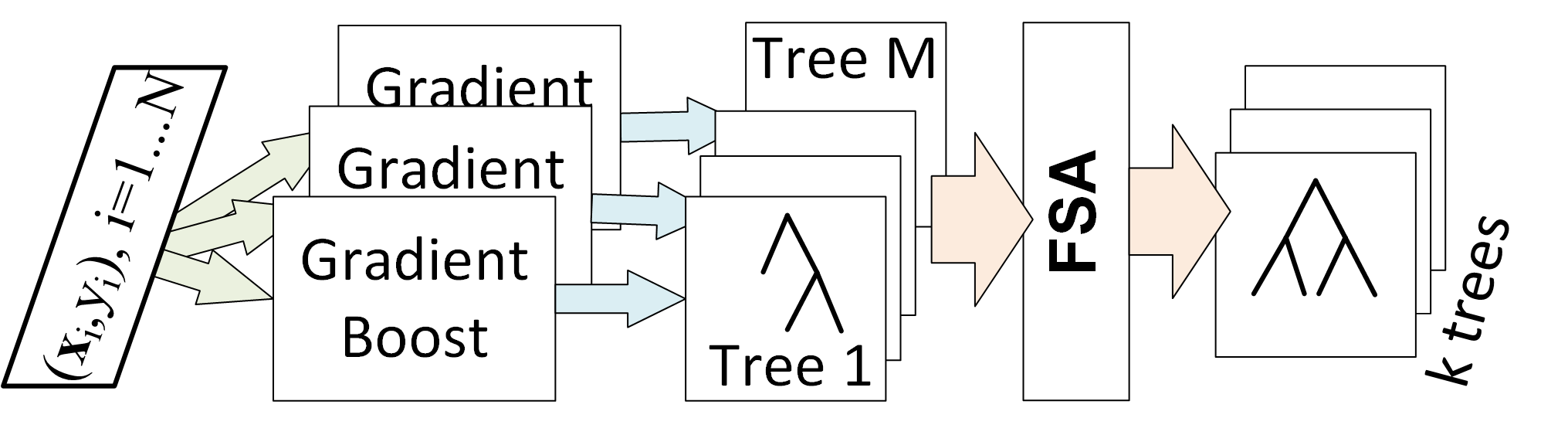}
\vspace{-3mm}
\caption{Diagram of the Relevant Ensemble of Trees algorithm.}\label{fig:ret}
\vspace{-2mm}
\end{figure*} 

The contributions of the paper are the following. 
\begin{enumerate}
\item It presents a novel way to obtain compact tree ensembles by generating a larger pool by GradientBoost and selecting a small number of trees by FSA, which also updates their leaf weights by loss minimization. A $L_2$ regularization term is added to the loss function for improved generalization. This is in contrast to boosting where $L_2$ regularization is hard to use and is often expressed in terms of a learning rate. 
\item It introduces a method to obtain more diverse trees in the initial pool by initializing many GradientBoost threads with random vectors instead of a constant bias. Even though these trees cannot be used as they are for prediction, they can be used after their leaf weights have been updated by loss minimization.  In addition, we prescribe different tree depths for the different threads, since our proposed formulation can naturally handle trees of different depths. 
\item It also presents a modification of FSA that obtains a large number of models with different sparsity levels in the same run, with minimal extra computation.
\end{enumerate}

\subsection{Related Work}
Trees based ensembles are regarded as one of the best off-the-shelf procedures for both classification and regression tasks. Applied naively, such ensembles often require a large of number of trees for modern big data sets. In addition, the complexity of each tree further increases with the size of the dataset. This jeopardizes their usability in large scale practical problems where memory is limited. 

There has been a large amount of work on different versions of boosting, which can be used for generating tree ensembles. Different versions of boosting minimize different loss functions, starting from Adaboost \cite{freund1995desicion} for the exponential loss, Logitboost \cite{friedman2000additive} for the logistic loss, and Gradientboost \cite{friedman2001greedy} for any differentiable loss. Other examples include Floatboost \cite{li2004floatboost} and Robust Logitboost \cite{li2012robust}. 

To facilitate enhanced interpretability and overcome memory based limitations, the problem of obtaining compact tree ensembles has received much attention in the recent years. 
An improved interpretability was aimed in \cite{friedman2008predictive} by selecting optimal rule subsets from tree-ensembles. 
The classical cost-complexity pruning of individual trees was extended in \cite{geurts2000some} to combined pruning of ensembles. 
The tree-ensemble based model was reformulated in \cite{joly2012l1} as a linear model in terms of node indicator functions and a L1-norm regularization based approach (LASSO) was used to select a minimal subset of these indicator functions. 
All of these works focused on the simultaneous pruning of individual trees and large ensembles. 

Another line of work focuses on lossless compression of tree ensembles. In \cite{painsky2016compressing} a probabilistic model was used for the tree ensemble and was combined with a clustering algorithm to find a minimal set of models that provides a perfect reconstruction of the original ensemble. Methods in \cite{geurts2000some}, \cite{joly2012l1}, \cite{painsky2016compressing} were developed for ensembles based on bagging or Random Forests only and exploit the fact that each of the individual trees is independent and identically distributed random entity for a given training data set. However, our method uses several threads of randomly initialized boosted ensembles. And it is well known that the trees generated with boosting are more diverse and much less complex compared to trees from bagging or Random Forests based models. 

Boosting was also used in \cite{reyzin2011boosting} to first train an entire tree ensemble. For prediction on a new example, the original tree ensemble was subsampled using a distribution induced by the tree weights. Since sampling of hypotheses needs to be done for every test example, besides being slow, one needs to store the entire original ensemble. The methods in \cite{painsky2016compressing} and \cite{reyzin2011boosting} aim only to provide an accurate description of the original trees using fewer models while our goal is to even improve upon the prediction accuracy of the original ensemble.

Recently, there has been some work on updating the leaf weights of already trained trees. The leaf weights of an already trained Random Forest \cite{breiman2001random} have been updated using an Artificial Prediction Market in \cite{barbu2012introduction}, which performs maximum likelihood learning in some cases \cite{barbu2012introduction}.  However, the Artificial Prediction Market did not obtain a compact tree ensemble by selecting a smaller set of trees, and was limited in the types of loss functions that can be optimized. 

In  \cite{ren2015global} the leaf weights of an already trained Random Forest were updated to minimize a loss similar to equation \eqref{eq:loss}. The trees leaves are also pruned by merging adjacent leaves according to a $L_2$ significance criterion. This achieved a similar goal with our paper in obtaining diverse trees of different depths, however no effort was done to obtain a compact set of trees that is smaller than the original set. 

\section{Relevant Ensemble of Trees (RET)}
We will work on regression and binary classification problems, where we are given training examples $\{(\mathbf{x_i},y_i)\in {\mathbb R}^p\times {\mathbb R},i=1,...,N\}$ and we need to find a prediction function $f_{\boldsymbol{\beta}}:{\mathbb R}^p\to {\mathbb R}$ parameterized by a parameter vector $\boldsymbol{beta}$, such that $f_{\boldsymbol{\beta}}(\mathbf{x_i})$ agrees with $y_i$ as much as possible. For example, for linear models, the prediction function is $f_{\boldsymbol{\beta}}(\mathbf{x})=\boldsymbol{\beta}^T\mathbf{x}$ and $\boldsymbol{\beta}\in {\mathbb R}^p$.

The agreement between $f_{\boldsymbol{\beta}}(\mathbf{x})$ and $y$ on the training examples is measured by a loss function. 
\vspace{-1mm}
\begin{equation}
L(\boldsymbol{\beta})=\sum_{i=1}^N \ell(f_{\boldsymbol{\beta}}(\mathbf{x_i}),y_i)+s(\boldsymbol{\beta})\label{eq:loss}
\vspace{-1mm}
\end{equation} that should be minimized, where $s(\boldsymbol{\beta})$ is a penalty such as the shrinkage $s(\boldsymbol{\beta})=\rho \|\boldsymbol{\beta}\|^2$ which helps with generalization. 

The loss $\ell(u,y)$ depends on the problem. For regression, it could be the square loss $\ell(u,y)=(u-y)^2$. For binary classification (when $y\in \{-1,1\}$), it could be the logistic loss $\ell(u,y)=\log(1+\exp(-uy))$, the hinge loss $\ell(u,y)=\max(1-uy,0)$, or other loss functions.

\subsection{Representation as Tree Ensemble} \label{sec:trees}

One needs a formal language to represent the prediction function $f_{\boldsymbol{\beta}}(\mathbf{x})$ in a computer. The set of all functions that could be obtained by the chosen prediction function representation is called the \textit{hypothesis space}. Our hypothesis space is the space of all tree ensembles, described below.

\textbf{Decision tree representation.} 
The basic building block of our representation is a decision tree. A decision tree is a function $T:{\mathbb R}^p \to {\mathbb R}$ associated with a tree representation that partitions the input domain ${\mathbb R}^p$ into a number of disjoint regions. Each region corresponds to a tree leaf and is characterized by the unique sequence of decisions that were made at the tree nodes to reach that leaf starting from the root. We will assume that each decision is based on one variable $x_j, j\in \{1,...,p\}$ of the input vector $\mathbf{x}\in {\mathbb R}^p$ and a threshold $\tau_j$, but this assumption is only related to the algorithm we use for building the trees and can be easily relaxed to other algorithms.

Following \cite{ren2015global}, we decompose the decision tree into a leaf weight vector $\boldsymbol{\beta}=(\beta_1,...,\beta_l)^T, \beta_i \in {\mathbb R}$ and an index function $\mathbf{i}_T(\mathbf{x}):{\mathbb R}^p\to\{0,1\}^l$, which is a column vector of all zeros except a single $1$ at the index of the leaf reached by the observation $\mathbf{x}\in{\mathbb R}^p$. Here $l=2^d$ is the maximum number of leaf nodes of a tree of maximum depth $d$. Thus, we can write the decision tree as a dot product $T(\mathbf{x})=\boldsymbol{\beta}^T \mathbf{i}_T(\mathbf{x})$. 
We will overload the notation by calling a decision tree as both the function $T(\mathbf{x}):{\mathbb R}^p\to {\mathbb R}$ and its decomposition $T=(\boldsymbol{\beta},\mathbf{i}_T(\mathbf{x}))$.

\textbf{Tree ensembles.} We will work with prediction functions that are the sum of a number of decision trees $k$, which are represented by their leaf weight vectors $\boldsymbol{\beta}_j$ and index functions $\mathbf{i}_j(\mathbf{x})$. Then, for a given feature vector $\mathbf{x}\in {\mathbb R}^p$, the prediction function has the form
\vspace{-1mm}
\[
\vspace{-1mm}
f(\mathbf{x})= \sum_{j=1}^k \boldsymbol{\beta}_j^T \mathbf{i}_j(\mathbf{x}).
\] 
Such models are usually called {\em tree ensembles}. 

\subsection{Training}

We are interested in parsimonious tree ensembles that minimize the loss function \eqref{eq:loss} constrained to contain at most $k$ trees (tree sparsity) of any depth in a prescribed set $S$. The parameters $k$ and $S$ are problem specific and can be obtained by cross-validation or using an information criterion such as AIC/BIC.

To select better trees and concurrently update their leaf weights to work best together, we will generate a large number of trees  $M$  by boosting and then use the Feature Selection with Annealing (FSA) algorithm \cite{barbu2016feature} to minimize the loss \eqref{eq:loss} constrained to be based on $k$ trees, as illustrated in Figure \ref{fig:ret}. This way in the end, we will have $k$ trees selected and their leaf weights optimized to minimize the loss function \eqref{eq:loss}.  

The steps of generating trees by boosting and then selecting them by FSA will be described in the next three subsections.

\subsection{Generating a pool of trees by GradientBoost}\label{sec:gb}

The initial pool of $M$ trees will be constructed by GradientBoost. In our applications, $M$ will be usually on the order of $M=3000$.
We will explore three approaches to generating these $M$ trees: 
\begin{itemize}
\item {\bf Single Chain Single Depth (SCSD)} generation where the $M$ trees of the same depth are obtained in a single GradientBoost run with $M$ boosting iterations.
\item {\bf Multi Chain Single Depth (MCSD)} generation where the $M$ trees of the same depth are obtained (in parallel) by $m$ separate GradientBoost runs with different random initializations, each run with $M/m$ boosting iterations.
\item {\bf Multi Chain Multi Depth (MCMD)} generation where the $M$ trees of $|S|$ different possible maximum depths are obtained (in parallel) by $m$ separate GradientBoost runs with different initializations, each run with $M/m$ boosting iterations.
\end{itemize}

Conventionally, GradientBoost is initialized with a constant prediction function, obtained by minimizing the loss function in terms of this constant value, which is referred to as the \textit{bias}.

\textbf{Random initialization.} Since  we are primarily interested in the structure of decision trees, in the multi chain approach we initialize the GradientBoost procedure with a random prediction vector. This allows us to invoke several versions of GradientBoost with different random initializations, resulting into a richer and more diverse collection of trees related to the underlying problem. 

Observe that the trees obtained from random initializations cannot be directly used for prediction without modifying their leaf weights, since they were started from random values (predictions) on the training examples, which make no sense on the test set. Through this random initialization we are only interested in obtaining the structure or mathematically speaking, index function $i(\mathbf{x})$ for each tree since our algorithm will update the leaf weights by loss minimization.

\textbf{What is a random prediction vector?} It refers to a vector containing the initial score or bias for every training instance. This base margin/score corresponding to every training instance is independently and identically sampled from N(0,1).  This randomization is performed only once and from there on, conventional GradientBoost is employed to obtain an ensemble of trees. Since bias is different for different training examples, one does not know what value of base score to use on a new, unseen test instance. As such, there is a need to establish a constant global bias for every example, which further necessitates the modification of leaf weights for all the trees in an ensemble. Our proposed algorithm “FSA on leaves” remedies all these issues while reducing tree redundancy and greediness to some extent. We consider it to be a novel extension to FSA algorithm, which was primarily developed for linear models.
 
\textbf{Why use a randomized base margin?} Let us talk about the first boosting iteration only. For a given loss function, the base margin of a training instance completely determines the gradient and Hessian entries corresponding to that instance. For example, using binomial deviance loss, positive instances (true label of 1) with base margins ln(1/9) ($<0$) and ln(9) ($>0$) have negative gradients as 0.9 and 0.1 respectively. For the first boosting iteration, in the case of a constant base margin for all the examples, all the positive instances have one and the same value for the gradient and such is the case for all the negative instances. However, for a randomized base margin, all the instances, positive or negative, have different gradient and Hessian values. Tree construction at every iteration depends only on the gradient and Hessian statistics. Instances with larger negative gradients have higher contribution in selecting best splitting variable and the best split point for a node. Roughly speaking, thus randomized base margin establishes a ranking among the examples in the order of their relevance towards tree building, unlike in constant base margin where all the positive instances are treated alike and so are all the negative instances. Different randomizations correspond to different rankings of the training examples, which results in a different tree at the first iteration. Since successive trees build on the previous trees, we expect the effects of randomization to percolate further deep down the iteration process. The motivation behind all this is to diversify the pool of trees as much as possible and have \textbf{FSA on leaves} pick up the best ones for the task. Observe that this is different from the compressed Random Forest \cite{joly2012l1} where all the trees are independently generated.  In contrast, our trees in each chain are dependent on each other because they are obtained by boosting.

Besides being faster, we will see empirically in Section \ref{sec:exp1} that the multi chain tree generation results in lower loss values and a more robust algorithm compared to the single chain tree generation. 

\textbf{Multiple depths.} In the case of Multi Chain Multi Depth (MCMD), we generate trees of different maximum depths given by the set $S$, obtaining a tree ensemble where the trees have a large range of depths. This is important because different features have different levels of interactions, which can be best captured with the correct tree depth. A smaller tree depth might not be sufficient  for fitting the interaction properly, while a larger depth might be overfitting. Empirical evidences in Section \ref{sec:exp1} validates the superiority of MCMD over the other tree generation approaches.

\subsection{Overview of the FSA algorithm}

We will use the Feature Selection with Annealing (FSA) algorithm \cite{barbu2016feature} to select the most relevant trees and update their weights.

FSA is an algorithm for simultaneous feature selection and model learning on the selected features, aimed at minimizing a differentiable loss function $L(\boldsymbol{\beta})$ with constraints on the number $k$ of non-zero coefficients:
\vspace{-1mm}
\begin{equation}
\boldsymbol{\beta}=\argmin_{|\boldsymbol{\beta}|_0\leq k} L(\boldsymbol{\beta}) \label{eq:fsa}
\end{equation}
\vspace{-1mm}

The FSA algorithm proceeds in a backward elimination manner, starting with $\boldsymbol{\beta}=0$ and alternating one stochastic gradient update step with a step that removes some variables according to a {\em deterministic} schedule that specifies the number $M_e$ of variables that should be left after iteration $e$. The FSA method is summarized in Algorithm \ref{alg:csfsa}.
 \begin{algorithm}[htb]
   \caption{{\bf Feature Selection with Annealing (FSA)}}
   \label{alg:csfsa}
\begin{algorithmic}
   \STATE {\bfseries Input:} Normalized training set $\{(\mathbf{x_i},y_i)\in {\mathbb R}^p\times {\mathbb R}\}_{i=1}^{N}$
   \STATE {\bfseries Output:} Trained model $f_{\boldsymbol{\beta}}(\mathbf{x})=\boldsymbol{\beta}^T\mathbf{x}$ with parameter vector $\mathbf{beta}$.
\end{algorithmic}
\begin{algorithmic} [1]
\STATE Initialize $\boldsymbol{\beta}=0$.
        \FOR {e = 1 to $N^{iter}$}
                \STATE  Update $\boldsymbol{\beta} \leftarrow \boldsymbol{\beta}-\eta \frac{\partial L(\boldsymbol{\beta})}{\partial \boldsymbol{\beta}}$
                \STATE Keep only the $M_e$ variables corresponding to the highest $|\boldsymbol{\beta}_j|$.
      \ENDFOR
\end{algorithmic}
\end{algorithm}
\vspace{-0mm}

The schedule $M_e, e=1,...,N^{iter}$ is quickly decreasing as 
\vspace{-1mm}
\[
\vspace{-1mm}
M_e=k+(p-k)\max(0,\frac{N^{iter}- 2 e}{2  e \mu +N^{iter}}),
\] 
specified by a parameter $\mu$, where $p$ is the dimension of the feature vectors $\mathbf{x_i}\in {\mathbb R}^p$.

Because most of the variables are removed in the first few iterations, the algorithm becomes increasingly fast after each iteration and can be hundreds of times faster than boosting when selecting thousands of variables. It is also thousands of times faster than the L1, SCAD and MCP penalized methods \cite{donoho2005stable},\cite{fan2001variable},\cite{zhang2010nearly}.

Besides being fast, another exciting fact about FSA is that it enjoys {\em theoretical guarantees of consistency and convergence} \cite{barbu2016feature}. If the learning rate is sufficiently small and the variable removal schedule is sufficiently slow, the FSA algorithm will find all the $k^*$ true variables $(k^*\leq k)$ with high probability. 
 
\subsection{Tree selection and leaf weight update by FSA}\label{sec:treesel}

In this section, we describe how to select a small number of trees $k$ from the pool of $M$ trees generated as described in Section \ref{sec:gb}.
Following the notation of Section \ref{sec:trees}, let these $M$ trees be $(\boldsymbol{\beta}_j,\mathbf{i}_j),j=\overline{1,M}$, where $\boldsymbol{\beta}_j$ is the leaf weight vector and $\mathbf{i}_j(\mathbf{x})$ is the index function of tree $j$.

We present two ways to select $k$ trees from the $M$ generated trees:

\noindent In {\bf FSA on trees}, the $M$ tree responses $T_j(\mathbf{x})=\boldsymbol{\beta}_j^T\mathbf{i}_j(\mathbf{x})$ are used by FSA as an $M$-dimensional feature vector to select $k$ features, thus the $k$ corresponding trees. If the sparse vector obtained by FSA training  is $\mathbf{w}=(w_1,...,w_M)\in{\mathbb R}^M$, then the obtained prediction function is 
\vspace{-1mm}
\[
\vspace{-1mm}
f_\mathbf{w}(\mathbf{x})=\sum_{j=1}^M w_j \boldsymbol{\beta}_j^T\mathbf{i}_j(\mathbf{x})
\]   
which has only at most $k$ non-zero coefficients $w_j$, thus depends on at most $k$ trees. 

Observe that FSA on trees can only be used for single chain tree generation since it doesn't update the leaf weights and the multi chain tree generation obtains randomly initialized trees that are not useful for prediction unless their leaf weights are updated.

\noindent In {\bf FSA on leaves}, we modify the FSA algorithm to update the tree leaf weights and select trees using a group criterion. For that, all the tree leaf weights are collected in a $l\times M$ matrix $B=(\boldsymbol{\beta}_1,\boldsymbol{\beta}_2,...,\boldsymbol{\beta}_M)$ of parameters for the prediction function
\vspace{-1mm}
\[
\vspace{-1mm}
f_B(\mathbf{x})=\sum_{j=1}^M \boldsymbol{\beta}_j^T\mathbf{i}_j(\mathbf{x}),
\]   
where $l$ is the maximum number of leaf nodes of the trees from our pool.

A variant of the FSA algorithm is run to minimize the loss function $L(B)$ from \eqref{eq:loss} with the matrix $B$ taking the place of $\boldsymbol{\beta}$. For that, the criterion for selecting the variables in step $4$ of FSA is changed to a group criterion such as $\|\boldsymbol{\beta}_j\|_2$ which ensures that only at most $k$ vectors $\boldsymbol{\beta}_j$ will be non-zero in the end. To correct for the bias towards selecting trees of larger depths, we further modify the group criterion for tree $j$ as $\frac{\|\boldsymbol{\beta}_j\|_2}{n_j}$, where $n_j$ is the number of leaves of tree $j$. 

\subsection{FSA with Many Sparsity Levels} \label{sec:skwarm}

For computational efficiency, we modify the FSA algorithm to obtain parameters for multiple sparsity levels by memorizing the obtained parameters at each iteration after a given starting iteration $N^0$ and optimizing them in separate routine.
The details are given in Algorithm \ref{alg:sk1} below.
\begin{algorithm}[htb]
   \caption{{\bf FSA with multiple sparsity levels}}
   \label{alg:sk1}
\begin{algorithmic}
   \STATE {\bfseries Input:} Normalized training examples $\{(\mathbf{x_i},y_i)\}_{i=1}^N$, learning rate $\eta$, sparsity levels $\{k_1,...,k_q\}, k_1>k_2>...>k_q$, number of iterations $N^{iter}$, annealing schedule $M_e,e=1,...,N^{iter}$.
   \STATE {\bfseries Output:} Trained classifier parameter vectors $\boldsymbol{\beta}_1,...,\boldsymbol{\beta}_q$ with $\|\boldsymbol{\beta}_i\|_0=k_i, i=1,...,q$.
\end{algorithmic}
\begin{algorithmic} [1]
\STATE Compute $E=\{e_1,...e_q\}$ with $e_i=\max \{e, M_e\geq k_i\}$.
\STATE Initialize $\boldsymbol{\beta}=0$.
        \FOR {$e=1$ to $N^{iter}$}
                \STATE  Update $\boldsymbol{\beta} \leftarrow \boldsymbol{\beta}-\eta \frac{\partial L(\boldsymbol{\beta})}{\partial \boldsymbol{\beta}}$
                \STATE Keep only the $M_e$ variables with highest $|\boldsymbol{\beta}_j|$ and renumber them $1,...,M_e$.
                \IF {$e\in E$}
                  \FOR {$i\in \{1,...,q\}$ such that $e_i=e$}
               \STATE Set $\boldsymbol{\beta}_i=\boldsymbol{\beta}$
                \STATE Keep only the $k_i$ variables $j$ with highest $|\boldsymbol{\beta}_{ij}|$ and renumber them $1,...,k_i$.
                \STATE  Update $\boldsymbol{\beta}_i \leftarrow \boldsymbol{\beta}_i-\eta \frac{\partial L(\boldsymbol{\beta}_i)}{\partial \boldsymbol{\beta}}$ for $N^{iter}$ times
      \ENDFOR
                \ENDIF
      \ENDFOR
\end{algorithmic}
\end{algorithm}

The annealing parameter $\mu$ in the interval $[10,20]$ works well in practice and FSA parameters are chosen in accordance with it.

In this paper we will use a modified version of this algorithm that has a matrix of leaf weights for $\boldsymbol{\beta}$ and a group criterion for selection, as described in Section \ref{sec:treesel}.

\section{Experiments}

We will perform three types of experiments. Our first experiment would ascertain the effectiveness of FSA in selecting relevant trees from a larger pool. The second set of experiments will compare the loss minimization obtained by GradientBoost with the ones obtained using RET with the three tree generation approaches, for the same model complexity. A smaller loss means a smaller training error which for the same model complexity would in general reflect in a lower test error unless the model overfits. The third set of experiments will compare the test error of the proposed method with GradientBoost (GB), XGBoost (XGB) and some linear  methods such as $L_1$ penalized logistic regression and Elastic Net on six real datasets.

\subsection{Simulation}\label{exp3}
\begin{figure}[htb]
\centering
\includegraphics[width=4.4cm]{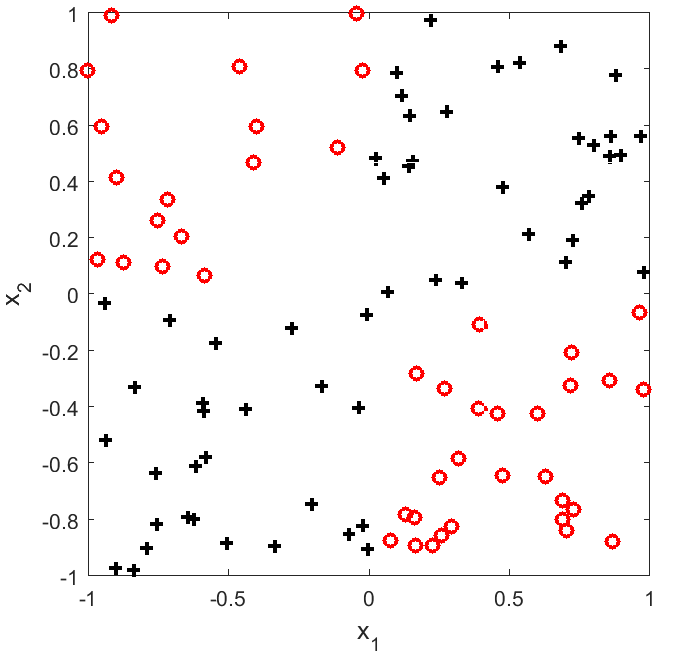}
\includegraphics[width=4.3cm]{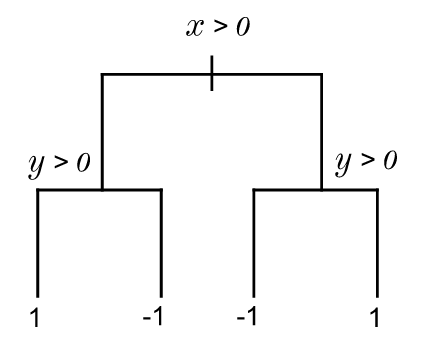}
\caption{An instance of the XOR dataset (left) and the decision tree that was used to generate the XOR data (right).}\label{fig:exp4}
\end{figure}

In this section, we use a classical example, the XOR data, to support our claim on the effectiveness of RET in obtaining compact tree ensembles. 
Since XOR, a non-linearly separable dataset, can perfectly be represented using a single decision tree of depth $2$ as illustrated in Figure \ref{fig:exp4}, right, we restrict the maximum tree depth to 2 for this experiment. Both the training and the test set consist of $100$ randomly sampled data points.

\noindent The setup of the experiment is as follows:
\begin{itemize}

\item \textbf{RET :} We use Single Chain Single Depth (SCSD) pool generation approach as described in \ref{sec:gb} to obtain an initial pool of $M=400$ trees of depth $2$. We then invoke \textit{FSA on leaves} as detailed in  \ref{sec:treesel} to select just one tree. 

\item \textbf{GradientBoost :}  We run GradientBoost procedure for the least number of iterations required to achieve about the same Test AUC as one given by RET. 
\end{itemize}

 \begin{table}[htb]
\vspace{-3mm}
\begin{center}
\caption{Simulated Experiment on XOR dataset, averaged over 100 runs.}\label{tab:exp2}
\begin{tabular}{lccc}
\hline 
\multicolumn{4}{l}{XOR dataset, $N=100, p=2, d=2$}\\ 
\hline \abovespace \belowspace
Method & \# trees $k$ &Train AUC &Test AUC\\
\hline \abovespace
GradientBoost &26 &\textbf{0.991} &0.967\\
\belowspace
RET  &1 &{0.985} &\textbf{0.968}\\
\hline
\end{tabular}
\end{center}
\vspace{-3mm}
\end{table}

In Table \ref{tab:exp2} are shown the area under an ROC curve for both training and the test sets, averaged over 100 independent runs. It takes GradientBoost, on an average, about 26 trees to match the generalization performance of a single tree given by RET.
From Table \ref{tab:exp2}, it is also apparent that the tree picked up by RET closely resembles the actual tree representation that was used for generating the  XOR data. This provides an empirical justification in support of the proposed Relevant Ensemble of Trees method using FSA for obtaining parsimonious tree ensembles.

 \begin{table*}[htb]
\caption{Dataset summary and RET/FSA parameter settings used in the experiments.}
\label{tab:data}
\centering
\begin{tabular}{lccc||ccccr}
\hline \abovespace \belowspace
	Dataset & obs train/test & features & classes & $\mu$&$\eta$ & $N^{iter}$ &$k_{max}$ & $S$(depth set)\\
\hline
\hline \abovespace
	gisette\cite{guyon2005result} &6,000/1,000 &5,000 & 2& 10&$10^{-3}$ & 300&600& $\{2,..,7\}$\Tstrut\\
	miniboone &130,065 &50 & 2&10&$10^{-3}$  & 300&600& $\{2,..,7\}$\Tstrut\\
	madelon\cite{guyon2005result} &2,000/600 &500 & 2&10& $10^{-3}$  & 150&100& $\{9,..,14\}$\Tstrut\\

	wilt &4,889 &6 & 2& 10&$10^{-3}$  & 150&100& $\{2,..,7\}$\Tstrut\\
	abalone &4,177 &8 & regression & 10&$10^{-5}$  & 150&100& $\{2,..,7\}$\Tstrut\\
\belowspace
	online news &39,797 &61 & regression & 10&$10^{-5}$  & 150& 100  & $\{2,..,7\}$\Tstrut\\
\hline
\hline
\end{tabular}
\end{table*}

 \begin{figure*}[t]
\centering
\includegraphics[width=5.7cm]{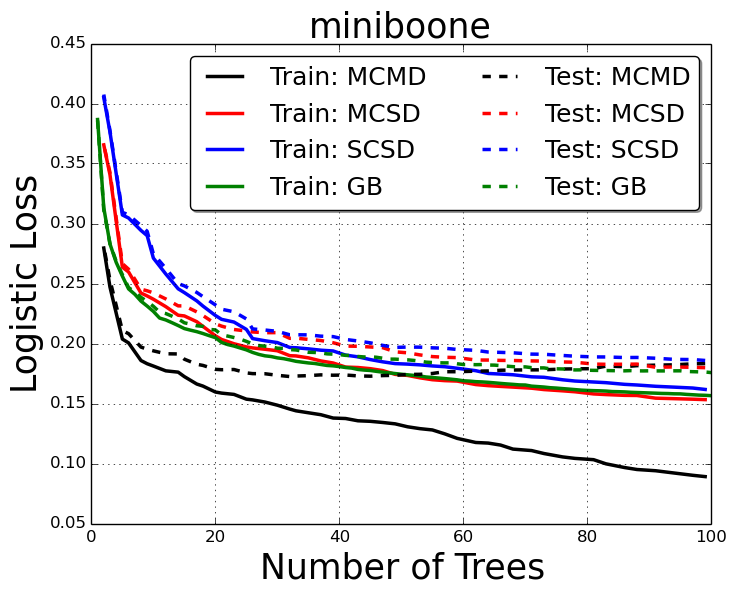}
\includegraphics[width=5.7cm]{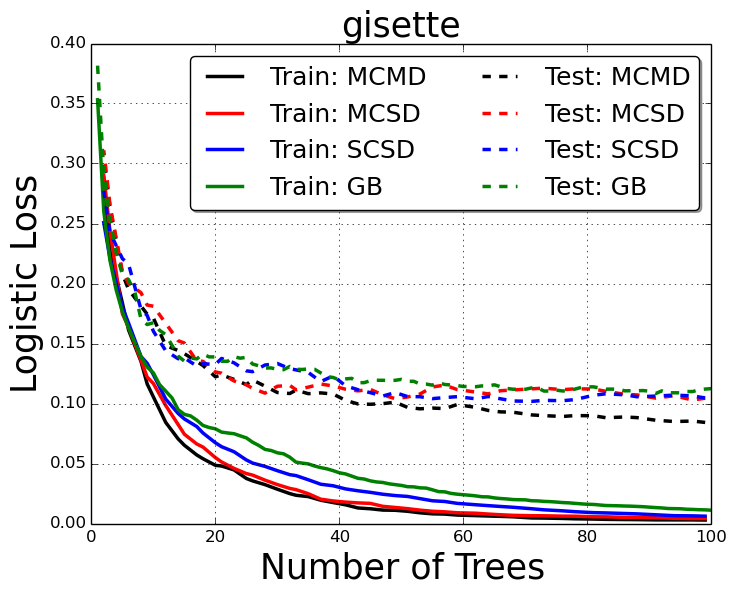}
\includegraphics[width=5.7cm]{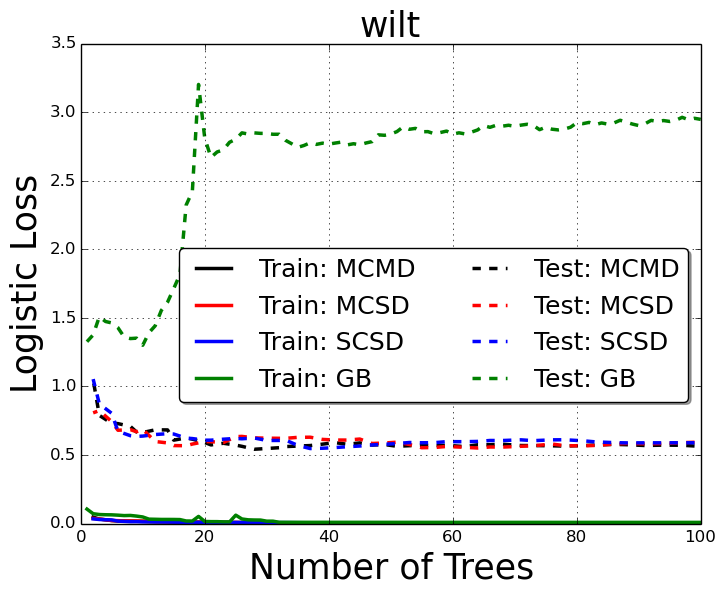}
\includegraphics[width=5.7cm]{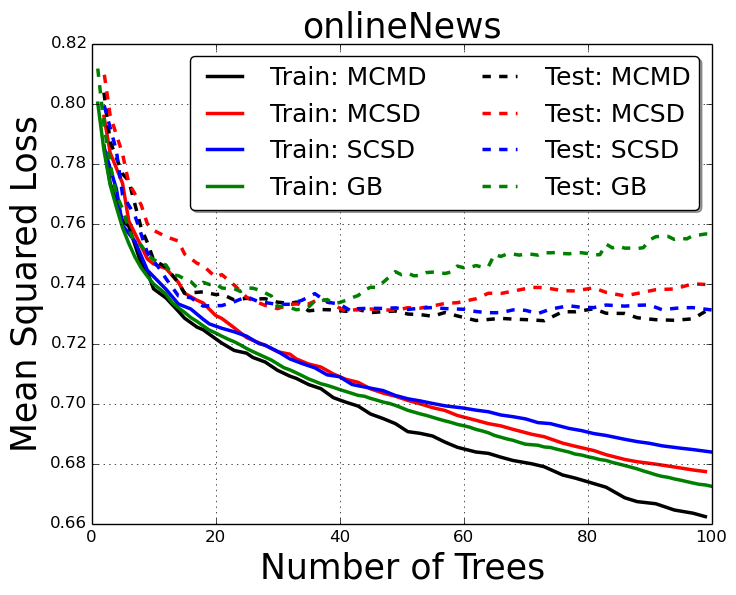}
\includegraphics[width=5.7cm]{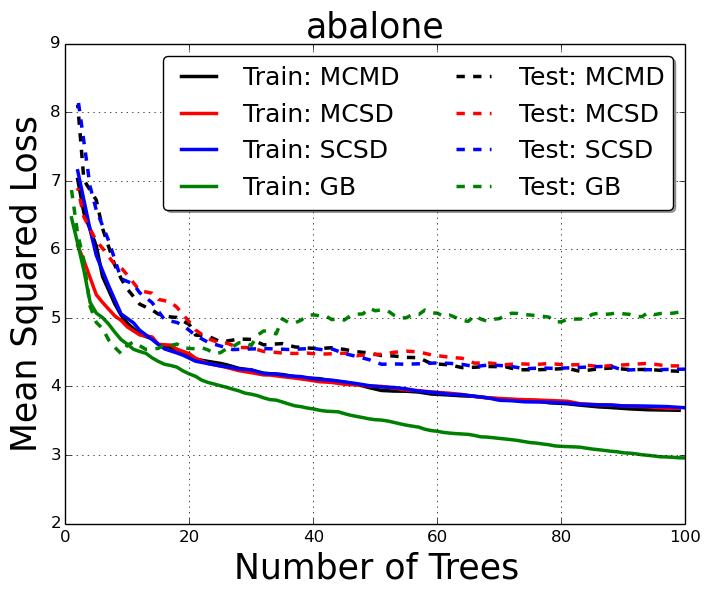}
\includegraphics[width=5.7cm]{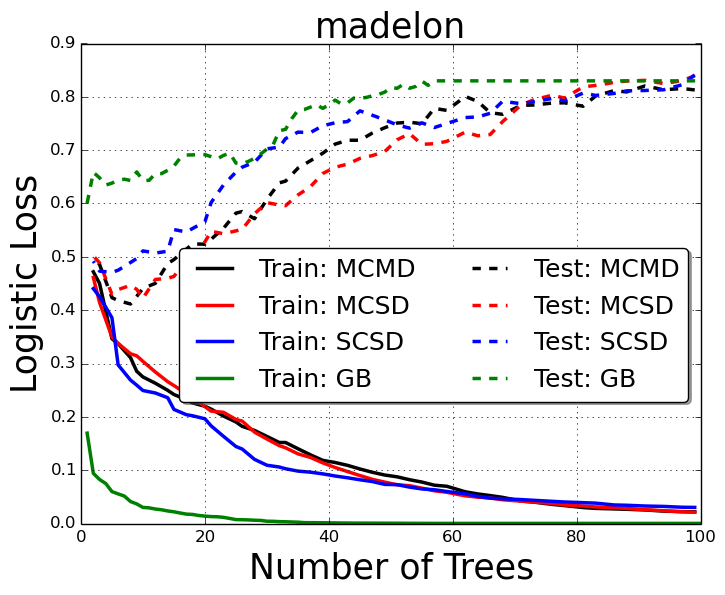}
\caption{Comparison of loss minimization for selecting different numbers of trees from pools generated using three different approaches and GradeintBoost.}\label{fig:exp1}
\vspace{-3mm}
\end{figure*}

\subsection{Datasets}\label{sec:data}

The experiments will be performed on six public datasets from the UCI machine learning repository \cite{asuncion2007uci}, of which $2$ have been part of the 2003 Feature Selection Challenge \cite{guyon2005result}. Since the FS challenge submission website is down, for the FS challenge datasets we have used the validation sets as test sets.
The datasets are summarized in Table \ref{tab:data}. 
From the datasets, we have removed any observations that had missing data.

\subsection{Loss Minimization Evaluation}\label{sec:exp1}


\begin{figure*}[!htb]
\centering
\includegraphics[width=5.cm]{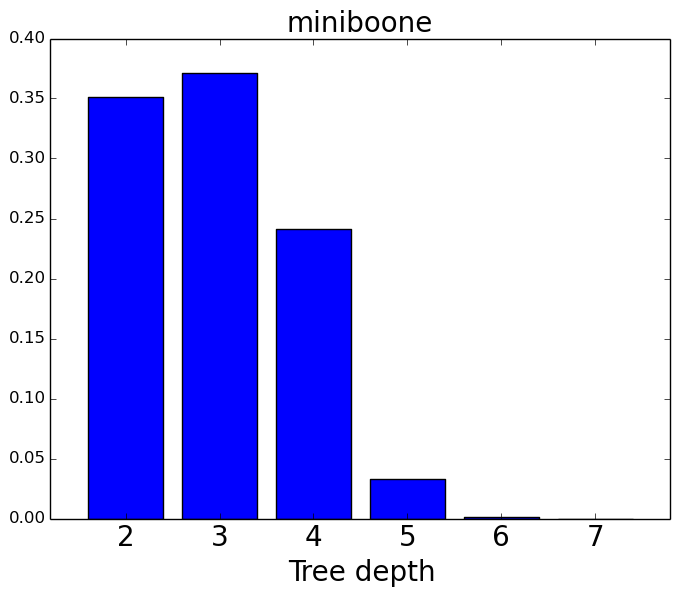}
\includegraphics[width=5.cm]{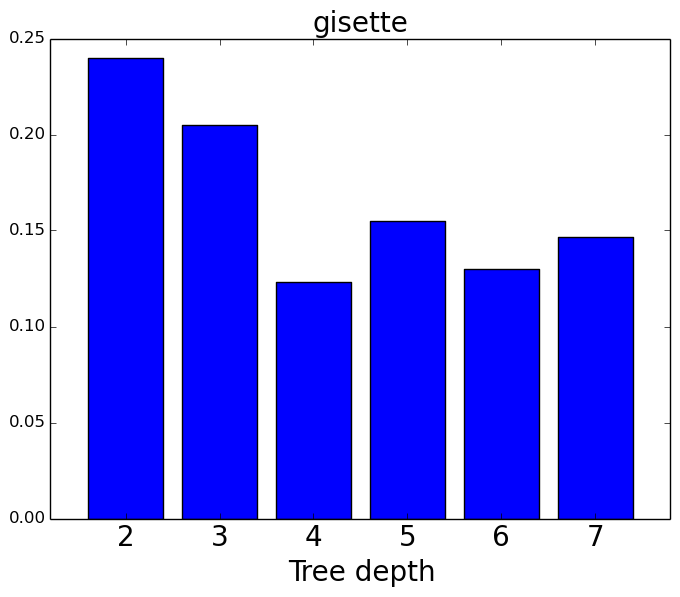}
\includegraphics[width=5.cm]{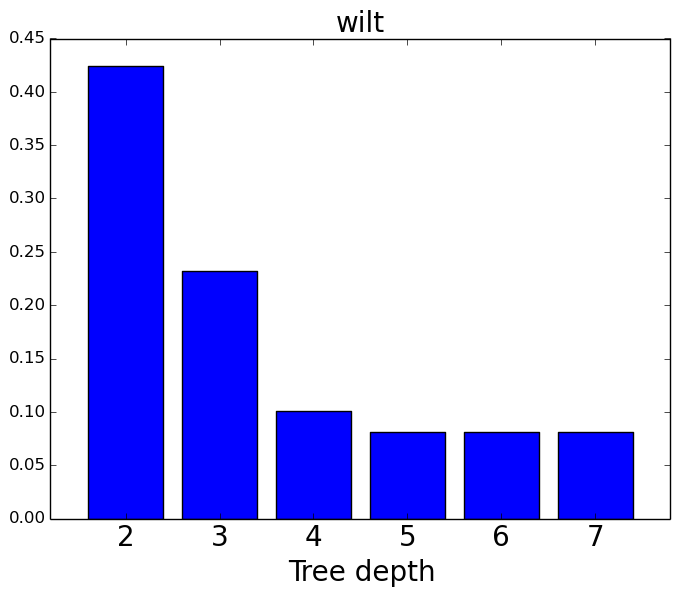}
\includegraphics[width=5.cm]{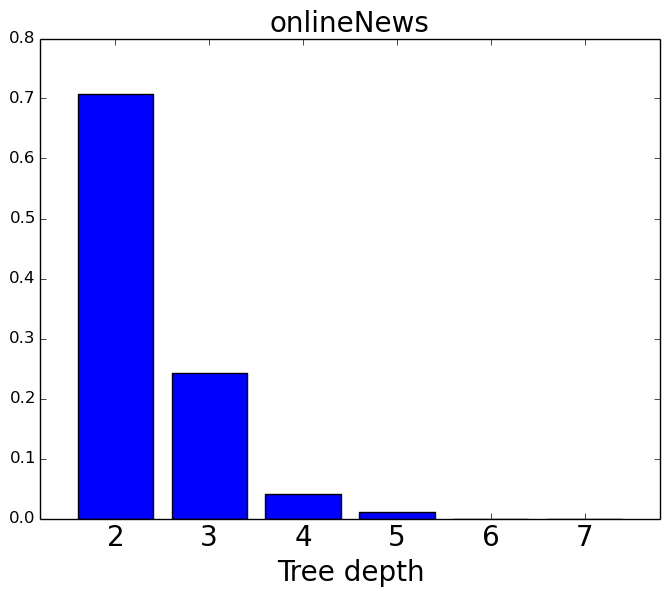}
\includegraphics[width=5.cm]{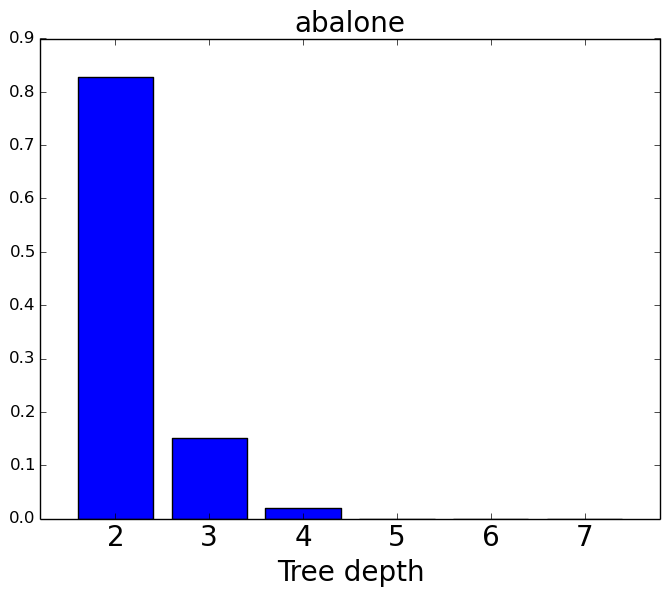}
\includegraphics[width=5.cm]{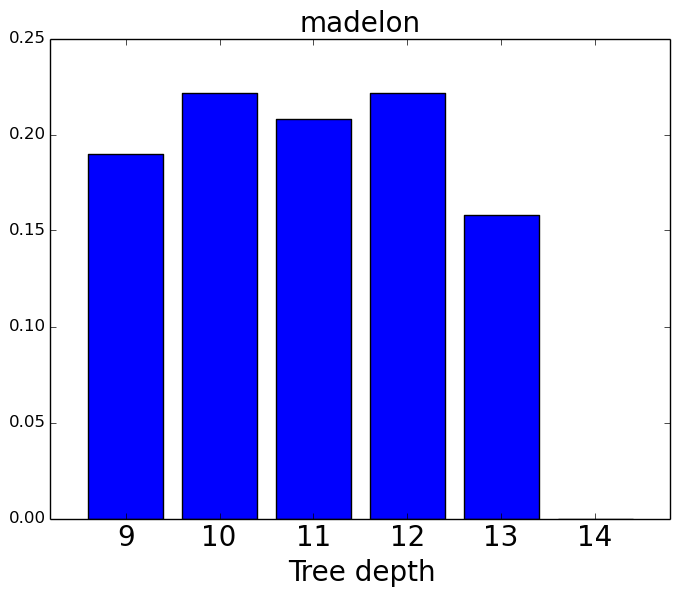}
\caption{Distribution of the fraction of trees selected by FSA  with tree depth.}\label{fig:exp2}
\end{figure*}

In this section, we will compare the loss minimization capabilities of the RET with GradientBoost in selecting the same number of trees. We will use pools of trees generated as follows:

\begin{itemize}
\item  \textbf{Single Chain Single Depth (SCSD)} pool generation: We use 3000 boosting iterations to obtain $M$=$3000$ trees of one single depth, which is obtained using the parameter tuning method described in detail in Section \ref{sec:testerror}.
\item \textbf{Multi Chain Single Depth (MCSD)} pool generation: We use 100 boosting iterations for each of the 30 chains to obtain $M$=$3000$ trees of the same depth as SCSD.
\item  \textbf{Multi Chain Multi Depth (MCMD)} pool generation: There are 30 chains, 5 chains for each depth in the depth set $S$, and 100 boosting iterations in each chain, for a total of $M$=$3000$ trees.
\end{itemize}
Other relevant parameters are given in Table \ref{tab:data}. For a fair comparison, the $L_2$ regularization parameter for RET was set to $\rho=0$.

The loss functions on the six datasets for the Gradient Boost and RET with the three pool generation methods (SCSD, MCSD, MCMD) are shown in Figure \ref{fig:exp1}. We see that the RET with the three pool generation methods generally overfit less than GradientBoost. Indeed, for all the six datasets, given a fixed number of trees, the RET methods usually have a lower test loss compared to GradientBoost  resulting in a more compact or relevant ensemble of trees.
Among different pool generation approaches, MCMD dominates on all but one (madelon) datasets as evidenced by test loss curves in Figure \ref{fig:exp1}, making it a natural choice for the desired  task. 

\begin{table*}[htb]
\centering
\caption{Real data results, averaged over 20 runs. Standard deviations are shown in parentheses.}\label{tab:errors}
\vskip 0.15in
\begin{tabular}{lcccccc}
\multicolumn{7}{c}{Classification datasets, test errors in \%.}\Tstrut\\
\hline
\abovespace \belowspace
Dataset &FSA &L1 &EL  &GB &XGB& RET\Tstrut\\
\hline
\abovespace
\textbf{gisette} &2.00 (0.14)  &3.80 (0.00) &3.77 (0.05) &4.78 (0.56)& 3.88 (0.39) &{\bf 1.95 (0.19)}\Tstrut\\
\textbf{madelon}&50.00 (0.00) &49.58 (0.43)  &49.37 (1.70) &23.3 (0.84) & 21.9 (1.13)&{\bf 17.6 (1.42)}\Tstrut\\
\textbf{miniboone}&12.22 (1.23) &9.84 (0.16) &9.83 (0.19) &6.70 (0.13) & 6.37 (0.06) &{\bf 5.64 (0.25)}\Tstrut\\
\belowspace
\textbf{wilt}&37.4 (0.00) &37.4 (0.00)  &37.4 (0.00) &26.09 (5.60)& 18.79 (1.60) &{\bf 17.14 (0.95)}\Tstrut\\
\hline 

\multicolumn{7}{c}{Regression datasets, test $R^2$ in \%.}\Tstrut\\
\hline
\abovespace
\textbf{abalone}&49.79 (2.98) &50.26 (3.10)  &50.27 (3.12) &50.14 (5.28)&51.98 (4.83) &{\bf 57.73(4.49)}\Tstrut\\
\belowspace
\textbf{online news}&11.94 (0.68) &10.61 (3.80)  &9.49 (5.17) &15.24 (0.24) & 14.43 (0.84)&{\bf 16.94 (0.80)}\Tstrut\\
\hline
\end{tabular}
\vspace{-3mm}
\end{table*}

\noindent {\bf Histograms.} For Multi Chain Multi Depth (MCMD) tree generation, the initial pool contains equal number of trees from each chain and equal number of trees for each depth. FSA is adept at automatically selecting trees corresponding to different tree depths using the group criterion specified in Section \ref{sec:treesel}. Figure \ref{fig:exp2} shows the proportion of trees selected by FSA from every depth type for six real datasets. 

We see that most of the tree depths are used in most datasets and that the distribution of tree depths differs from dataset to dataset.

\subsection{Test Error Evaluation}\label{sec:testerror}

In this section, we compare RET with GradientBoost (GB), XGBoost (XGB), some generalized linear models with the $L_1$ \cite{article},\cite{donoho2005stable},\cite{zhao2006model} and Elastic Net \cite{zou2005regularization} penalties, as well as linear FSA to see if we really need tree-based models. 
For a more accurate comparison, we will use the logistic loss  $\ell(u,y)=\log(1+\exp(-uy))$ for linear FSA and RET.

{\bf Parameter tuning}. The parameters for the methods being evaluated have been obtained by five-fold cross-validation on the training set as follows:
\begin{enumerate}
\item For each parameter combination, the cross-validated loss was computed  as the average of the validation loss over the five cross-validation folds of the training set.
\item The parameter combination corresponding to the smallest cross-validated loss was selected and the final model was obtained for that parameter combination by training on the entire training set. This model was then used to obtain the predictions on the test set.
\end{enumerate}

The parameters involved in parameter tuning are the following:
\begin{itemize}
\item The number of selected trees $k\in [1, k_{max}]$ for RET and for GradientBoost/XGBoost (as $k$ boosting iterations). We used 50 sparsity levels $k$ on an exponential grid between $1$ and $k_{max}$.
\item The tree depth $d$ for GradientBoost. It has the range  $d\in \{2,...,7\}$ for most datasets and $d\in\{9,...,14\}$ for madelon.
\item The shrinkage parameter $\rho$ for FSA, Elastic Net and RET, $\rho\in$ $\{10^{-1}, 10^{-2}, ..., 10^{-4}, 10^{-5}\}$ 
\end{itemize}

Other parameters for FSA and RET were fixed to values given in Table \ref{tab:data}. 
The test errors for the datasets without a test set are obtained using a random $80-20$ train-test split. All results are shown as averages over 20 independent runs.
%


 In Table \ref{tab:errors} are shown the test errors using the parameter tuning described above. We see that RET obtains the lowest test errors on all of the four binary classification tasks and the highest test $R^2$ on both regression datasets. 
 \begin{table}[htb]
\centering
\caption{Average training times in minutes.}\label{tab:times}
\vskip 0.15in
\begin{tabular}{lccccccc}
\hline
\abovespace \belowspace
Dataset &FSA &L1 &EL &GB &XGB & RET\Tstrut\\
\hline
\abovespace
\textbf{gisette}  &16.7 &42 &257 &112 & 16.9&535\Tstrut\\
\textbf{madelon} &6.4 &1.7 &8.3 &3.7 & 0.6 &225 \Tstrut\\
\textbf{miniboone}&4.6 &101 &439  &248 &107&857\Tstrut\\
\textbf{wilt}&0.02 &0.29 &1.4  &0.67 &0.69&6.1\Tstrut\\
\textbf{abalone}&0.07 &0.09 &0.45  &0.13 & 0.09&7.2\Tstrut\\
\belowspace
\textbf{online news}&16.2 &46  &208 &23 &0.42  &85\Tstrut\\
\hline
\end{tabular}
\end{table}

 One downside of RET is that it has higher training times as shown in Table \ref{tab:times}, which can partly be improved upon exploiting the parallelization of MCMD pool generation  process. The FSA model building for the different parameter combinations on the different cross-validation folds can also be easily parallelized.

\section{Conclusion}

This paper presents a novel method for obtaining a compact tree ensemble based on loss minimization that involves generating a large pool of trees followed by selecting a desired number of trees while updating their leaf weights using the FSA algorithm. The initial pool of trees is obtained by Boosting, thus it is more compact and relevant than the trees obtained by Random Forest, leading to a more parsimonious tree ensemble.
Experiments on six UCI datasets indicate that the proposed approach usually obtains a smaller test loss for the same model  complexity than GradientBoost and smaller test errors on large datasets.

One of the novel ideas of this paper is to generate the initial pool of trees using many parallel GradientBoost (GB) threads having random initializations and different tree depths. This approach obtains more diverse trees than a single GB thread or even compared to many GB threads trained on bootstrap samples. The leaf weights of these trees cannot be used for prediction and had to be retrained by loss minimization using the FSA algorithm. The usage of $L_2$ regularization in RET makes it more robust to overfitting compared to GradientBoost. 

Another contribution is the modification in FSA algorithm that obtains a whole range of models corresponding to different sparsity levels in just a single run. 
The obtained sets of weights have the property that each contains the next one, thus forming a chain relative to the inclusion relation. This is an FSA equivalent to using warm restarts in the Elastic Net.

\nocite{pedregosa2011scikit}
\bibliographystyle{IEEEtran}
\bibliography{IEEEexample}

\end{document}